\pdfoutput=1
\documentclass[11pt]{article}
\usepackage[]{acl}
\usepackage{times}
\usepackage{latexsym}
\usepackage[T1]{fontenc}
\usepackage[utf8]{inputenc}
\usepackage{microtype}
\usepackage{inconsolata}
\usepackage{booktabs}
\usepackage{graphicx}
\usepackage{graphics}
\usepackage{enumitem}
\usepackage{dcolumn}

\newlist{questions}{enumerate}{2}
\setlist[questions,1]{label=RQ\arabic*.,ref=RQ\arabic*}
\setlist[questions,2]{label=(\alph*),ref=\thequestionsi(\alph*)}

\author{Jenny Kunz\thanks{Equal Contribution} \\
Linköping University \\
  \texttt{jenny.kunz@liu.se} \\\And
  Oskar Holmström\footnotemark[1] \\
  Linköping University \\
  \texttt{oskar.holmstrom@liu.se} \\}

\title{The Impact of Language Adapters in Cross-Lingual Transfer for NLU}

\begin{document}
\maketitle
\begin{abstract}
Modular deep learning has been proposed for the efficient adaption of pre-trained models to new tasks, domains and languages. In particular, combining language adapters with task adapters has shown potential where no supervised data exists for a language. 
In this paper, we explore the role of language adapters in zero-shot cross-lingual transfer for natural language understanding (NLU) benchmarks. We study the effect of including a target-language adapter in detailed ablation studies with two multilingual models and three multilingual datasets. 
Our results show that the effect of target-language adapters is highly inconsistent across tasks, languages and models. Retaining the source-language adapter instead often leads to an equivalent, and sometimes to a better, performance. Removing the language adapter after training has only a weak negative effect, indicating that the language adapters do not have a strong impact on the predictions. 
\end{abstract}

\section{Introduction}
\label{sec:intro}

Adding smaller components to a large language model (LLM) that can be specifically targeted, trained, stacked and exchanged is becoming increasingly common \citep{pfeiffer2023modular}. Particularly adapters \citep{pmlr-v97-houlsby19a} and LoRA \citep{hu2021lora} are widespread for the efficient adaption of LLMs. They often perform on par or better than fine-tuning the models' parameters while avoiding issues of interference such as catastrophic forgetting \citep{mccloskey1989catastrophic, ratcliff1990connectionist}.
 
In this work, we focus on pre-trained target-language adapters for zero-shot cross-lingual transfer. \citet{pfeiffer-etal-2020-mad} found that any cross-lingual transfer problem can be decomposed in language and task, and introduce a setup that combines task and language adapters, both independently trained on top of a pre-trained multilingual model. This setup is appealing particularly for low-resource and medium-resource languages that lack high-quality data for supervised training as it can be applied to unseen task-language combinations. However, how consistent the effect of the target-language adapter is has not been explored explicitly. In particular, it has not been explored how including target-language adapters compares to keeping the source-language adapter for the cross-lingual transfer. 
In addition, the detailed ablations by \citet{pfeiffer-etal-2020-mad} focus on named entity recognition, while it remains unclear if similar results also hold for higher-level language understanding tasks. Therefore, we focus on three multilingual natural language understanding (NLU) benchmarks.
We investigate the following questions:
\begin{questions}
    \item \textit{How robust is the positive effect of adding a target-language adapter across languages, models and tasks?} To answer this question, we compare the performance with target-language adapters to other setups that keep the source-language adapter or that only include task adapters. \label{rq1}
    \item \textit{How much does the model rely on the effect of the language adapters?} We test this with a setup that leaves out the language adapter without substitution, and measure the performance drop. \label{rq2}
    \item \textit{Does the amount of source-language and target-language pre-training data in the base model affect the effect of the target-language adapter?} We compare the effect of target-language and source-language adapters conditioned on the languages' representation in the pre-training corpora. \label{rq3}
\end{questions}

Surprisingly, our extensive ablations show that instead of using the target-language adapter, we can often retain the source-language adapter that was used during training, or even leave out the language adapter after training with no negative (or even positive) effects on the models' performance. Even a setup that does not include language adapters at all is competitive and sometimes better. The results are however inconsistent across models, datasets and language pairs. We observe a higher benefit of target-language adapters for lower-resource target languages, but only for one out of four model-task combinations. 

We conclude that the contribution of language adapters is less clear than we thought and that they do not play an interpretable role in the decision-making for language understanding tasks. However, they sometimes have a strong positive effect on the performance, making it worthwhile to test them in scenarios where they could be useful. We suggest putting more effort into understanding if there are interpretable properties of the base model, task, source language or target language that cause gains when using language adapters.

\section{Related Work}
\label{sec:rel_work}

\paragraph{Modular Deep Learning. }Modular deep learning has gained attention with the primary goal of adapting pre-trained models to new tasks and languages efficiently, but also to avoid issues of interference such as catastrophic forgetting \citep{mccloskey1989catastrophic, ratcliff1990connectionist} and the curse of multilinguality \citep{conneau-etal-2020-unsupervised}. 
Adapters \citep{pmlr-v97-houlsby19a} introduce a small number of additional parameters, which increases the inference overhead \citep{hu2021lora} but shows promising performance. For large-enough models (>3B parameters), language-specific adapters are even reported to outperform continued pre-training on unseen target languages \citep{yong2022bloom}. On the other hand, \citet{ebrahimi-kann-2021-adapt} report that for the XLM-R \citep{conneau-etal-2020-unsupervised} model, language adapters perform inferior to target-language fine-tuning. 
Crucially, post-hoc fine-tuning of adapters reportedly performs on par with including them in pre-training \citep{kim-etal-2021-revisiting}, which makes them particularly attractive where computational resources are limited. 

\paragraph{Language Adapters.} For language transfer with adapters, some work has focused on aggregating information from related languages, language families and genera. In the study by \citet{lauscher-etal-2020-zero}, syntactic tasks rely heavily on language similarity, while it is less pronounced (though still existent) for semantic tasks. 
The UDapter framework \citep{ustun-etal-2020-udapter} integrates language adapters in a syntactic dependency parsing model, conditioned on typological features of the language. 
\citet{faisal-anastasopoulos-2022-phylogeny} adapt MLMs to unseen languages using hierarchical adapters inspired by phylogenetic trees. The tree hierarchy enables linguistically informed parameter sharing between related languages, leading to strong performance gains, especially for very low-resource languages and zero-shot transfer. 
This structured approach is apparently getting more consistent results than continued pre-training, where a diverse set of languages can top related languages \citep{fujinuma-etal-2022-match}. 

The MAD-X framework \citep{pfeiffer-etal-2020-mad} combines independently trained language and task adapters. Input embeddings are also processed by \emph{invertible adapters}, whose inverse processes the output embeddings. They report successful cross-lingual transfer even for unseen combinations, making it possible to use models even where no annotated data exists for a language and even if the language was unseen during model pre-training. For cross-lingual transfer from a \emph{monolingual} model, \citep{artetxe-etal-2020-cross}'s results indicate some improvement using Houlsby-style language adapters over exchanging the token embeddings only for NLU tasks . However, \citet{ebrahimi-kann-2021-adapt} report that for languages unseen during pre-training, performing continued pre-training outperforms training language adapters and invertible adapters. \citet{he-etal-2021-effectiveness} explore task adapters (with no language adapters) for cross-lingual transfer on XLM-R and find that they perform better than fine-tuning, both on the full data and on low-resource setups. They hypothesize that adapters better maintain the target-language knowledge from pre-training as the original model's parameters are not changed. 
\citet{pfeiffer-etal-2022-lifting} propose a framework that introduces language modularity at pre-training time, overcoming interference at no parametric cost. 


\section{Experimental Setup}
\label{sec:exp_setup}

In the following, we introduce the models, adapters, adapter training setups, ablation setups and datasets that we use for our ablation studies of language adapters.  A link to our code including hyperparameters used to run our experiments will be published after the anonymity period.  The code, including the hyperparameters used to run our experiments, is available at \url{https://github.com/oskarholmstrom/lang-adapters-impact}.

\subsection{Model and Adapters}
We use XLM-Roberta-base (XLM-R), trained on 100 languages \citep{xlm, conneau-etal-2020-unsupervised}, and multilingual BERT (mBERT), trained on 104 languages \citep{devlin-etal-2019-bert}. Most languages we test on are included in the pre-training of both models with the exception of Haitian Creole (ht) for XLM-R and Quechua (qu) for both models.
We use pre-trained language adapters from AdapterHub \citep{pfeiffer-etal-2020-adapterhub}. We train task-specific Pfeiffer adapters using AdapterHub's associated \emph{adapter-transformers} library\footnote{\url{https://github.com/adapter-hub/adapter-transformers}}. Only task adapter parameters and classification heads are trained; language adapters and model parameters are kept frozen. 

\paragraph{Adapter Setups. } We train models with source-language adapters and evaluate them on the target language in three setups: 
\begin{itemize}
    \item \emph{Target} replaces source-language adapters with target-language adapters at evaluation time.
    \item \emph{Source} keeps the source-language adapters even at evaluation time.
    \item \emph{None} leaves out the language adapter entirely at evaluation time (although still trained with source-language adapters).
\end{itemize}

\noindent To test if language adapters are beneficial at all, we include a fourth setup:
\begin{itemize}
    \item In \emph{None}$_{tr}$, models are both trained and evaluated without language adapters. Only task adapters are included in the models. 
\end{itemize}

\paragraph{Pre-Training Data. }

For ablations that test the effect of the representation of the source- and target language in the pre-training corpus, we create a ranking. For XLM-R, we use the data on language representation given in the original paper \citep{xlm}. mBERT is trained on Wikipedia data\footnote{Source: \url{https://github.com/google-research/bert/blob/master/multilingual.md}}. While no exact numbers or details on the dump are given, we estimate the size with the current number of articles for each language\footnote{\url{https://meta.wikimedia.org/wiki/List_of_Wikipedias} (version: 2023/12/15)}. Wikipedia data was also used for the pre-training of the language adapters.

\begin{table}[ht]
\centering
\begin{tabular}{lrr}
\toprule
Lang.\ & XLM-R \small{(\#Tokens)} & mBERT \small{(\#Articles)} \\ \midrule
Ar & 2,869M & 1.2M \\
De & 10,297M & 2.9M \\
El & 4,285M & 229K\\
En & 55,608M & 6.8M \\
Es & 9,374M & 1.9M \\
Et & 843M & 241K \\
Hi & 1,715M & 160K \\
Ht & not included & 69K \\
Id & 2,2704M & 676K \\
Ja & 530M & 1.4M \\
Qu & not included &  not included (24K) \\
Ru & 23,408M & 2.0M \\
Sw & 275M & 79K \\
Tr & 2,736M & 543K\\
Vi & 24,757M & 1,3M \\
Zh & 259M+176M & 1.4M \\ \bottomrule
\end{tabular}
\caption{Representation of languages in the pre-training corpora of the models. The mBERT data is approximated with the current number of Wikipedia articles. Quechua was not included in mBERT's pre-training. Wikipedia data was also used for the pre-training of the language adapters. }
\label{tab:pre-train-langs}
\end{table}

\subsection{Data Sets}

We evaluate language adapters on three natural language understanding and commonsense reasoning data sets. All data sets include human translations from the English original into several diverse languages, and are balanced with respect to the different labels. 
XCOPA is the only of the three data sets that was also included in the original MAD-X evaluation \citep{pfeiffer-etal-2020-mad}.

\paragraph{PAWS-X. } 
English PAWS \citep{zhang-etal-2019-paws} is a paraphrase detection data set. Specifically, the task is to classify if a pair of sentences is a paraphrase or not. PAWS includes 108,463 paraphrase and non-paraphrase pairs deliberately chosen to have a high lexical overlap. 
PAWS-X \citep{yang-etal-2019-paws} is a multilingual extension of English PAWS. It includes 51401 examples human-translated into German (de), Spanish (es), French (fr), Japanese (ja), Korean (ko) and Chinese (zh). 

\paragraph{XNLI. } The Multi-Genre Natural Language Inference (MultiNLI) corpus \citep{williams-etal-2018-broad} is a multi-genre corpus with the goal of classifying the entailment relation of a pair of sentences. Possible labels are \textit{entailment}, \textit{neutral} or \textit{contradiction}. The corpus contains a total of 432,702 sentence pairs.
XNLI \citep{conneau-etal-2018-xnli} extends MultiNLI with human translations into Arabic (ar), Bulgarian (bg), German (de), Greek (el), Spanish (es), French (fr), Hindi (hi), Russian (ru), Swahili (sw), Thai (th), Turkish (tr), Urdu (ur), Vietnamese (vi) and Chinese (zh).

\paragraph{XCOPA. } The Choice Of Plausible Alternatives (COPA) dataset \citep{roemmele2011choice, gordon-etal-2012-semeval} is part of the SuperGLUE benchmark \citep{superglue} and consists of 500 training and 500 test examples. Each example consists of a premise, a question (\textit{What was the CAUSE?} or\textit{ What happened as a RESULT?}) and two answer options. The task is to select the option that is more likely to have a causal relation with the premise. 
XCOPA \citep{ponti-etal-2020-xcopa} is a multilingual extension that includes human translations of the \emph{evaluation} data into Estonian (et), Haitian Creole (ht), Indonesian (id), Italian (it), Eastern Apurímac Quechua (qu), Kiswahili (sw), Tamil (ta), Thai (th), Turkish (tr), Vietnamese (vi), and Mandarin Chinese (zh). 

\subsection{Evaluation Setup}

For each experiment, we report the mean accuracy over five random seeds. For better comparability across models, we only include the languages from the data sets for which pre-trained language adapters exist on AdapterHub for both models.

\section{Results}
\label{sec:results}

Given the large number of combinations of models, tasks and language pairs in our experiments, we summarise them and present individual results of particular interest in this section. The full results can be found in Appendix \ref{appx:full_results}.

\subsection{General Trends}

Overall, as we see in table \ref{tab:models_all} that the \textit{None}$_{tr}$ model is the best-performing setup. For the individual models, there is however always a similar-performing setup that includes language adapters: For XLM-R, the \textit{Target} setup has the same performance, while for mBERT, the difference to \textit{Source} is negligible ($0.1\%$).
For XLM-R, using \textit{Target} has an advantage of $2.4\%$ over \textit{Source}, but for mBERT, it is vice versa with a difference of $2.1\%$. 

\begin{table}[ht]
\centering
\begin{tabular}{ c c c c c} 
\toprule
& Target & Source & None & None$_{tr}$  \\ \midrule
XLM-R & \textbf{72.6} & 70.2 & 71.0 & \textbf{72.6}  \\
mBERT & 62.7 & 64.8 & 59.8 & \textbf{64.9}  \\
\bottomrule
\end{tabular}
\caption{Average results for each model over all languages and datasets (XNLI, PAWS-X and XCOPA).}
\label{tab:models_all}
\end{table}

Breaking down the results by datasets, we see in table \ref{tab:all} that the best-performing setup varies notably. All setups except \textit{None} perform best for at least one model-task combination. And while \textit{None}$_{tr}$ was the best overall, we see that \textit{Target} performs the best on three out of six combinations. Note in this context that the results in table \ref{tab:models_all} were not adjusted for the number of languages included in the datasets, leading to the smaller PAWS-X set being underrepresented. The difference between \textit{Target} and \textit{None} varies from $0.6\%$ to $5.4\%$, showing that the reliance of the model on the language adapter is inconsistent.

\begin{table*}[t]
\centering
\begin{tabular}{ c c c c c c c c c} 
\toprule
\multicolumn{1}{c}{} & \multicolumn{4}{c}{XLM-R} &  \multicolumn{4}{c}{mBERT} \\  
& Target & Source & None & None$_{tr}$ & Target & Source & None & None$_{tr}$    \\  \cmidrule(lr){2-5} \cmidrule(lr){6-9}
XNLI & 72.1 & 69.4 & 70.3 & \textbf{72.4} & 60.5 & 62.9 & 57.9 & \textbf{63.3}   \\
PAWS-X & \textbf{80.9} & 80.1 & 80.3 & 80.8 & 76.7 & \textbf{78.0} & 71.3 & 77.0  \\
XCOPA & \textbf{53.7} & 51.9 & 52.3 & 50.3 & \textbf{52.3} & 51.3 & 51.4 & 51.4  \\
\bottomrule
\end{tabular}
\caption{Average results for all model-task combinations. }
\label{tab:all}
\end{table*}

\subsection{Transfer from English}

We now zoom into the different target languages, focusing on cross-lingual transfer with English as the source language. This is arguably the most realistic scenario due to the large amount of annotated data available in English. Similar tables for other source languages are presented in Appendix \ref{appx:full_results}.

\paragraph{PAWS-X. }

The results for PAWS-X are reported in table \ref{tab:paws-x_en}. For XLM-R, all setups show a relatively similar performance, with the range of the average across languages being between $77.3\%$ (\textit{English} and \textit{None}) and $78.2\%$ (\textit{None}$_{tr}$). For mBERT, \textit{None} is an outlier with a strong drop in performance that is consistent across all target languages, getting an accuracy of only $69.4\%$ instead of $76.3$-$77.4\%$, while keeping the English source-language adapter is the best setup in all languages. 

\begin{table*}[t]
\centering
\begin{tabular}{ c c c c c c c c c} 
\toprule
\multicolumn{1}{c}{} & \multicolumn{4}{c}{XLM-R} &  \multicolumn{4}{c}{mBERT} \\
& Target & English & None & None$_{tr}$ & Target & English & None & None$_{tr}$ \\ \cmidrule(lr){2-5} \cmidrule(lr){6-9}
En & (\textbf{91.4}) & (\textbf{91.4}) & (91.0) & (91.1) & (\textbf{91.3}) & (\textbf{91.3}) & (82.7) & (90.4) \\
De & \textbf{83.3} & 82.3 & 82.4 & 83.2 & 81.1 & \textbf{82.2} & 73.1 & 81.2 \\
Es & 84.0 & \textbf{84.1} & 83.5 & \textbf{84.1} & 82.0 & \textbf{83.1} & 72.8 & 81.6 \\
Ja & 69.7 & 69.2 & 69.6 & \textbf{70.2} & 69.7 & \textbf{69.9} & 64.1 & 69.1 \\
Zh & 74.3 & 73.7 & 73.8 & \textbf{75.1} & 72.6 & \textbf{73.6} & 67.8 & 73.4 \\
\midrule Avg. & 77.8 & 77.3 & 77.3 & \textbf{78.2} & 76.4 & \textbf{77.2} & 69.4 & 76.3 \\ \bottomrule
\end{tabular}
\caption{Results on PAWS-X with transfer from English (en) into all evaluated target languages, ordered by pre-training resources top-to-bottom. Results on English are included for reference but excluded from the average.}
\label{tab:paws-x_en}
\end{table*}

\paragraph{XNLI. }

Results for XNLI are reported in table \ref{tab:xnli_en}.
For XLM-R, the \textit{None}$_{tr}$ setup that is trained and evaluated without language adapters performs best, and this is the case for 7 out of 10 cross-lingual evaluation languages and for English. Comparing \textit{Target} and \textit{Source}, there is a small advantage for using the target-language adapters (on average $70.6$ versus $70.0\%$), but the results are inconsistent over target languages: For 5 evaluation languages, the target-language adapter is better, for 4 languages, the English adapter is better, and for one language, they get the same results.
For mBERT, keeping the English adapter is the overall best setup with $63.0\%$ (and the best for 9 out of 10 languages), followed by \textit{None}$_{tr}$ with 62.2\%. Exchanging the adapter and especially leaving it out after training can have a strong negative effect for mBERT, showing a higher reliance on the language adapter parameters: The drop when using \textit{None} as compared to using the English adapter that was active during training is $9.4$ percentage points. 

\begin{table*}[t]
\centering
\begin{tabular}{ c c c c c c c c c} 
\toprule
\multicolumn{1}{c}{} & \multicolumn{4}{c}{XLM-R} &  \multicolumn{4}{c}{mBERT} \\
& Target & English & None & None$_{tr}$ & Target & English & None & None$_{tr}$ \\ \cmidrule(lr){2-5} \cmidrule(lr){6-9}
En & (81.8) & (81.8) & (81.5) & (\textbf{81.7}) & (\textbf{78.1}) & (\textbf{78.1}) & (70.9) & (77.7) \\
De & \textbf{73.6} & 73.3 & 73.4 & \textbf{73.6} & 66.1 & \textbf{67.9} & 58.1 & 67.5 \\
Ru & 72.4 & 72.4 & 72.7 & \textbf{72.8} & 64.1 & \textbf{64.6} & 55.0 & 64.1 \\
Es & 76.0 & \textbf{76.2} & 75.9 & 75.9 & 69.1 & \textbf{71.4} & 62.5 & 70.5 \\
Zh & 70.0 & \textbf{71.7} & 70.8 & 71.0 & 66.3 & \textbf{67.4} & 57.7 & 65.8 \\
Vi & 71.6 & 71.5 & 71.3 & \textbf{71.8} & 68.2 & \textbf{68.4} & 58.7 & 66.8 \\
Ar & 68.6 & 65.8 & 68.2 & \textbf{68.8} & 38.7 & \textbf{62.7} & 50.7 & 61.9 \\
Tr & 69.8 & 70.7 & 70.2 & \textbf{71.0} & 62.0 & \textbf{61.3} & 50.6 & 60.4 \\
El & \textbf{72.3} & 71.9 & 71.8 & 72.0 & 60.8 & \textbf{60.9} & 54.0 & 60.2 \\
Hi & 66.7 & 67.1 & 66.9 & \textbf{67.2} & 57.1 & \textbf{57.4} & 47.6 & 56.5 \\
Sw & 65.2 & 59.0 & 62.4 & \textbf{62.7} & 37.4 & 47.7 & 40.8 & \textbf{48.2} \\
\midrule Avg. & 70.6 & 70.0 & 70.4 & \textbf{70.7} & 59.0 & \textbf{63.0} & 53.6 & 62.2 \\ \bottomrule
\end{tabular}
\caption{Results on XNLI with transfer from English (en) into all evaluated target languages, ordered by pre-training resources top-to-bottom. Results on English are included for reference but excluded from the average.}
\label{tab:xnli_en}
\end{table*}

\paragraph{XCOPA. }

Results for XCOPA are reported in table \ref{tab:xcopa_en}.
For XLM-R, target-language adapters increase the performance consistently compared to all other setups. \textit{None}$_{tr}$ is the lowest-performing setup by a notable margin ($50.3\%$ compared to $52.0$-$53.8\%$ for the other setups), showing that this model-task combination draws the strongest positive effect from including language adapters in the training.
The results for mBERT are more mixed: While \textit{Target} performs best on average, it only performs better than the English adapter for half of the languages. Compared to the other two datasets, exchanging adapters after training does not have a negative impact on mBERT; the English adapter is even the worst on average, while \textit{Target} is the best setup with a margin of $1.0$ to $1.1\%$. 

For XLM-R, there are previous results by \citet{pfeiffer-etal-2020-mad}. Our accuracy scores are lower than theirs. However, our results are not directly comparable to theirs as they perform sequential fine-tuning of the task adapter that additionally contains the SIQA dataset, what reportedly improves the performance on XCOPA \citep{sap-etal-2019-social}. 

\begin{table*}[t]
\centering
\begin{tabular}{ c c c c c c c c c} 
\toprule
\multicolumn{1}{c}{} & \multicolumn{4}{c}{XLM-R} &  \multicolumn{4}{c}{mBERT} \\
& Target & English & None & None$_{tr}$ & Target & English & None & None$_{tr}$ \\ \cmidrule(lr){2-5} \cmidrule(lr){6-9}
Zh & \textbf{55.2} & 55.0 & 54.3 & 49.4 & 53.7 & 52.7 & \textbf{54.2} & 53.2 \\
Vi & \textbf{55.3} & 54.9 & 55.1 & 52.8 & 51.6 & 52.9 & 51.1 & \textbf{52.6} \\
Tr & \textbf{53.1} & 51.9 & 51.2 & 49.3 & 51.9 & 53.2 & 54.1 & \textbf{55.6} \\
Id & \textbf{55.7} & 53.6 & 53.4 & 49.8 & 50.4 & \textbf{50.8} & \textbf{50.8} & \textbf{50.8} \\
Et & \textbf{54.1} & 50.7 & 52.3 & 51.4 & \textbf{53.8} & 49.3 & 49.1 & 51.2 \\
Sw & \textbf{54.0} & 49.7 & 52.0 & 49.7 & 50.0 & 50.4 & \textbf{50.5} & 49.1 \\
Ht & \textbf{51.2} & 48.6 & 50.6 & 49.6 & \textbf{54.6} & 52.7 & 51.2 & 50.2 \\
Qu & \textbf{51.4} & 51.2 & 49.6 & 50.2 & \textbf{52.6} & 48.5 & 49.8 & 48.2 \\
\midrule Avg. & \textbf{53.8} & 52.0 & 52.3 & 50.3 & \textbf{52.3} & 51.3 & 51.4 & 51.4 \\ \bottomrule
\end{tabular}
\caption{Results on XCOPA with transfer from English (en) into all evaluated target languages, ordered by pre-training resources top-to-bottom. }
\label{tab:xcopa_en}
\end{table*}

\subsection{Effect of Pre-Training Data}

In this section, we contrast the amount of pre-training data of source and target languages by visualising the improvement of using the target-language adapter as compared to keeping the source-language adapter. This is inspired by \citet{pfeiffer-etal-2020-mad}'s evaluation that finds that adding language adapters helps more for the transfer from high-resource to low-resource languages in named entity recognition.
Note that for XCOPA, training data only exists for English, therefore we limit this analysis to PAWS-X and XNLI. 

\paragraph{PAWS-X. }

\begin{figure}[htbp]
\centering
\includegraphics[width=0.49\linewidth]{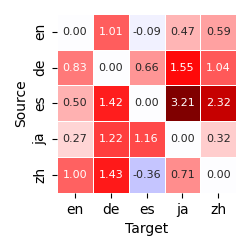}
\includegraphics[width=0.49\linewidth]{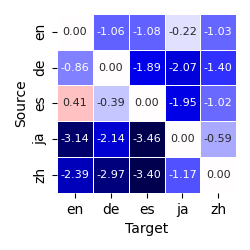}
\caption{Difference between the target-language adapter and source-language adapter on PAWS-X for XLM-R (left) and mBERT (right) for each source and target language. The amount of pre-training data decreases top-to-bottom/left-to-right.}
\label{fig:paws-x}
\end{figure}

The cross-lingual transfer for PAWS-X, as seen in Figure \ref{fig:paws-x}, does not show a consistent pattern. For mBERT, we see that having a lower-resource source language correlates with a decreased performance with the target-language adapter. It has to be noted though that for this dataset, none of the evaluated languages is particularly low-resource, as we can see in Table \ref{tab:pre-train-langs}. 

\paragraph{XNLI. }

\begin{figure*}[htbp]
\centering
\includegraphics[width=0.49\linewidth]{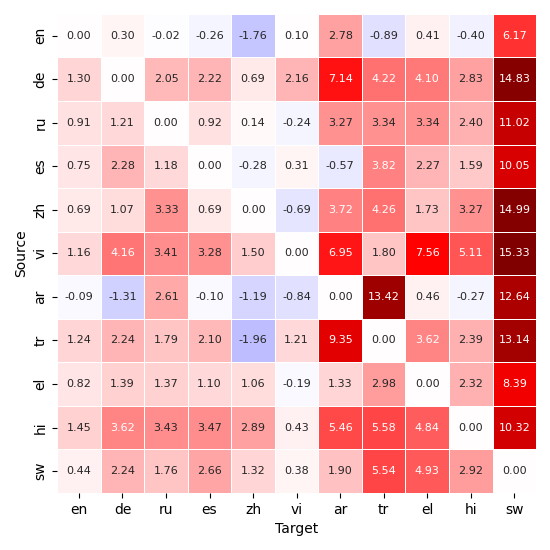}
\includegraphics[width=0.49\linewidth]{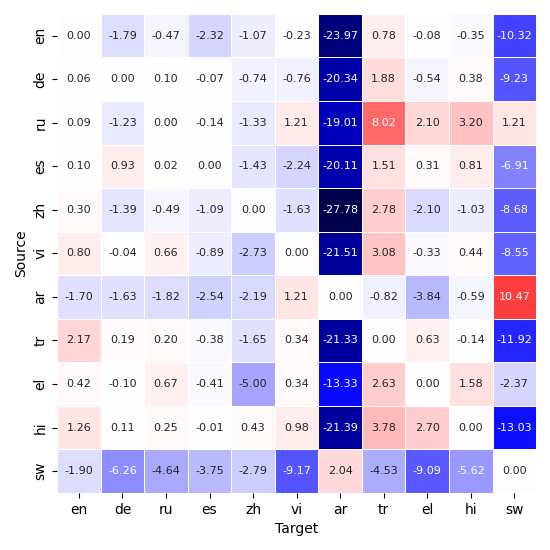}
\caption{Difference between the target-language adapter and source-language adapter on XNLI with XLM-R (left) and mBERT (right) for each source and target language. The amount of pre-training data decreases top-to-bottom/left-to-right.}
\label{fig:xnli_target-source}
\end{figure*}

For the XNLI data set, we report the results for both models in Figure \ref{fig:xnli_target-source}. For XLM-R, we observe a tendency for lower-resource target languages to benefit more, as the right side of the Figure has higher numbers. A strong outlier effect is visible for the lowest-resource language in our evaluation, Swahili, where the gains from the target-language adapter are bigger than for all other target languages by a large margin. Surprisingly, we also see that the benefit of \textit{Target} for English as a source language is smaller than for all other source languages.
For mBERT, we do not see a general pattern across all or most of the lower-resource languages. However, with Swahili and Arabic, two outliers show a strongly \textit{negative} effect from their target-language adapters, except when transferred to each other (and, for Swahili, from Russian).

\section{Discussion}

In Section \ref{sec:results} have observed relatively inconsistent results regarding the utility of language adapters, and of target-language adapters in particular. In the following, we discuss the relation of our results to the research questions introduced in Section \ref{sec:intro}, as well as the variance across datasets, limitations of our experiments, and avenues for future work.

\subsection{Effect of Target-Language Adapters (\ref{rq1})} 
\label{secsec:rq1}
The positive effect of adding a target-language adapter instead of keeping the source-language adapter is inconsistent. While the XLM-R model gains on average $2.4\%$ across all combinations of tasks, source languages and target languages, the mBERT model loses on average $2.1\%$ (Table \ref{tab:models_all}). For the XCOPA dataset, the target-language adapters appear to be crucial to transfer skills, especially for the XLM-R model but to a lesser extent also for mBERT. 
For the other two datasets, the results are however mixed. 
Even where the target-language adapter has an advantage, keeping the source-language adapter does not hurt the performance much. This indicates that while zero-shot cross-lingual transfer is possible, for the languages we test on, the performance does not rely much on the target-language adapters. It also indicates that we do not observe a strong isolated modular effect of the language adapters. In line with previous results by \citet{he-etal-2021-effectiveness}, we hypothesise that much of the target language performance comes from the frozen base model's multilingual capabilities, combined with the task adapter and classification head.
This is also confirmed by the finding that no language adapter at all (the \textit{None}$_{tr}$ setup) often performs on par or better than the models with language adapters. 

\subsection{Reliance on Language Adapters (\ref{rq2})}
\label{secsec:rq2}
The drop in performance when removing the language adapter that was included at training time without substitution is weak for XLM-R which loses only $1.6\%$ compared to the \textit{Target} setup and $0.8\%$ compared to the \textit{Source} setup. For mBERT however, it is much stronger, with $-2.9\%$ compared to the \textit{Target} and $-5.0\%$ compared to the \textit{Source} setup. mBERT appears to be more sensitive to adapter changes after training, indicating that it relies more on the parameters of the language adapters than the relatively robust XLM-R model. However, it does not appear that the language adapter parameters themselves are heavily important, as \textit{None}$_{tr}$ does not see a similar drop. We conclude that the contribution of the language adapters is small.

Related results indicating that the modular role of adapters is inconsistent and not always predictable have been reported by \citet{ruckle-etal-2021-adapterdrop} pruning adapters from AdapterFusion models to reduce inference time. They show that this is often possible without sacrificing task performance.  

\subsection{Effect of Pre-Training Resources (\ref{rq3})}
\label{secsec:rq3}
We do not observe a consistent pattern that would indicate that transfer from high-resource to lower-resource languages is more beneficial. In this respect, the NLU benchmarks appear to differ from named entity recognition, where \citet{pfeiffer-etal-2020-mad} observed a strong effect. That lower-resource languages benefit more is notable for the combination of the XLM-R model and XNLI, but not for the other three model-task combinations. For source languages, we do not see the expected effect; on the contrary, English as the source language has the \textit{worst} record for \textit{Target}. We do however note large differences between language pairs, and outlier languages that benefit or lose more than other languages. This suggests that while language adapters and specifically target-language adapters are not always beneficial, it is worthwhile to test them for every target language individually. 

Looking at Quechua, which is not included in the pre-training of either model, and Haitian Creole, which is not included in the pre-training of XLM-R, we observe a positive effect of the target-language adapter. However, both languages are included only in the XCOPA dataset which benefits most from target-language adapters in general, and do not stand out with a higher margin to the \textit{Source} setup than other languages.  

\subsection{Variance across Datasets}
We have observed that for XCOPA, the target-language adapters are more crucial, while for PAWS-X and XNLI, the cross-lingual transfer works similarly well without the language adapter, based on the multilingual capabilities of the pre-trained base model only. 
A natural question arising from this observation is what causes these differences. One obvious fact is that COPA is a harder task, with models reaching a relatively low performance. In comparison, XNLI is translated from MultiNLI which is reportedly robust to random word-order permutations \citep{sinha-etal-2021-unnatural}, indicating that lexical cues and less nuanced interactions between words play a large role. This is confirmed by the results of \citet{kew2023turning} who compare English versus multilingual instruction fine-tuning of LLMs for cross-lingual transfer and find that for highly structured tasks like XNLI, the language of the fine-tuning plays less of a role.
To what extent this is also the case for COPA examples that the models succeed on remains to be tested. 

Another hypothesis is that the translations play a role. The translations of XCOPA may be less close to the English source, making a better command of the target language crucial. Closer and more literal translations of PAWS-X and XNLI may enable an easier inheritance of skills learned in English.

\subsection{Limitations and Future Work}
\paragraph{Architecture. }While we do not observe higher increases from \textit{Source} to \textit{Target} for lower-resource languages, there remain large differences in overall performance that correlate with pre-training resources, indicating that cross-lingual transfer is far from a solved problem. The potential of language adapters to narrow this gap has not been exhaustively tested in this work. We have only explored the Pfeiffer adapter architecture and only one single language adapter at a time. As we discussed in Section \ref{sec:rel_work}, there are alternative methods which can be explored. 
The analysis could even be extended with models introducing modularity already at pre-training time \citep{pfeiffer-etal-2022-lifting}, which has a different scope but may reveal important insights.

A factor that may limit the potential of language adapters trained post-hoc is the finding that cross-lingual capabilities  emerge late in pre-training, as reported by \citet{blevins-etal-2022-analyzing} doing probing studies on pre-training checkpoints of XLM-R. More work on the interactions of languages in multilingual models, and the prerequisites for successful cross-lingual transfer, may inform the design and training of language adapters in the future. 

\paragraph{Languages and Data. }Another avenue for future work is a more thorough investigation of adapters for more languages not included in the base model's pre-training. Even adapters for new languages in monolingual models \citep{artetxe-etal-2020-cross} would be an insightful addition to our analysis. 
A limiting factor, as in the present work, is the lack of high-quality language understanding benchmarks that cover a broad set of languages. In addition, all datasets we use are translations from the English original, which commonly introduces translation artefacts translation artifacts \citep{gellerstam1986translationese, freitag-etal-2019-ape}. 
The creation of more such datasets would enable a better understanding of cross-lingual transfer methods. 

\section{Conclusion}
In this work, we performed extensive ablations on cross-lingual transfer with pre-trained language adapters for NLU benchmarks. 
We found that the inclusion of target-language adapters appears to have a small benefit on average, but it is slight and varies significantly across languages, models and tasks. As the effect is not robust and we do not observe patterns clear enough to predict it, it remains to be tested for each use case and language individually. 
Keeping the source-language adapter often has a surprisingly good performance, and for one of two models, even leaving out the adapter without substitution is possible without large performance drops. This shows that the model does not rely much on the language adapter, and that language adapters do not appear to be an impactful isolated language module. 

While this work provides new insights into the utility of language adapters for NLU, many questions remain open. 
We conclude that there is a need to identify the specific conditions --- such as properties of the base model, task, source, and target languages --- under which language adapters enhance performance, and thereby unlocking their usefulness in a broader setting. 

\section*{Acknowledgments}

We thank the anonymous reviewers for their insightful and constructive feedback. 
The research in this paper was funded by the National Graduate School of Computer Science in Sweden (CUGS) and by the European Commission under grant agreement no. 101135671.
The computations were enabled by resources provided by the National Academic Infrastructure for Supercomputing in Sweden (NAISS) at Alvis partially funded by the Swedish Research Council and by the Berzelius re- sources provided by the Knut and Alice Wallenberg Foundation at the National Supercomputer Centre.

\bibliography{anthology,custom}

\begin{thebibliography}{35}
\expandafter\ifx\csname natexlab\endcsname\relax\def\natexlab#1{#1}\fi

\bibitem[{Artetxe et~al.(2020)Artetxe, Ruder, and Yogatama}]{artetxe-etal-2020-cross}
Mikel Artetxe, Sebastian Ruder, and Dani Yogatama. 2020.
\newblock \href {https://doi.org/10.18653/v1/2020.acl-main.421} {On the cross-lingual transferability of monolingual representations}.
\newblock In \emph{Proceedings of the 58th Annual Meeting of the Association for Computational Linguistics}, pages 4623--4637, Online. Association for Computational Linguistics.

\bibitem[{Blevins et~al.(2022)Blevins, Gonen, and Zettlemoyer}]{blevins-etal-2022-analyzing}
Terra Blevins, Hila Gonen, and Luke Zettlemoyer. 2022.
\newblock \href {https://aclanthology.org/2022.emnlp-main.234} {Analyzing the mono- and cross-lingual pretraining dynamics of multilingual language models}.
\newblock In \emph{Proceedings of the 2022 Conference on Empirical Methods in Natural Language Processing}, pages 3575--3590, Abu Dhabi, United Arab Emirates. Association for Computational Linguistics.

\bibitem[{Conneau et~al.(2020)Conneau, Khandelwal, Goyal, Chaudhary, Wenzek, Guzm{\'a}n, Grave, Ott, Zettlemoyer, and Stoyanov}]{conneau-etal-2020-unsupervised}
Alexis Conneau, Kartikay Khandelwal, Naman Goyal, Vishrav Chaudhary, Guillaume Wenzek, Francisco Guzm{\'a}n, Edouard Grave, Myle Ott, Luke Zettlemoyer, and Veselin Stoyanov. 2020.
\newblock \href {https://doi.org/10.18653/v1/2020.acl-main.747} {Unsupervised cross-lingual representation learning at scale}.
\newblock In \emph{Proceedings of the 58th Annual Meeting of the Association for Computational Linguistics}, pages 8440--8451, Online. Association for Computational Linguistics.

\bibitem[{Conneau and Lample(2019)}]{xlm}
Alexis Conneau and Guillaume Lample. 2019.
\newblock \href {https://proceedings.neurips.cc/paper/2019/file/c04c19c2c2474dbf5f7ac4372c5b9af1-Paper.pdf} {Cross-lingual language model pretraining}.
\newblock In \emph{Advances in Neural Information Processing Systems}, volume~32. Curran Associates, Inc.

\bibitem[{Conneau et~al.(2018)Conneau, Rinott, Lample, Williams, Bowman, Schwenk, and Stoyanov}]{conneau-etal-2018-xnli}
Alexis Conneau, Ruty Rinott, Guillaume Lample, Adina Williams, Samuel Bowman, Holger Schwenk, and Veselin Stoyanov. 2018.
\newblock \href {https://doi.org/10.18653/v1/D18-1269} {{XNLI}: Evaluating cross-lingual sentence representations}.
\newblock In \emph{Proceedings of the 2018 Conference on Empirical Methods in Natural Language Processing}, pages 2475--2485, Brussels, Belgium. Association for Computational Linguistics.

\bibitem[{Devlin et~al.(2019)Devlin, Chang, Lee, and Toutanova}]{devlin-etal-2019-bert}
Jacob Devlin, Ming-Wei Chang, Kenton Lee, and Kristina Toutanova. 2019.
\newblock \href {https://doi.org/10.18653/v1/N19-1423} {{BERT}: Pre-training of deep bidirectional transformers for language understanding}.
\newblock In \emph{Proceedings of the 2019 Conference of the North {A}merican Chapter of the Association for Computational Linguistics: Human Language Technologies, Volume 1 (Long and Short Papers)}, pages 4171--4186, Minneapolis, Minnesota. Association for Computational Linguistics.

\bibitem[{Ebrahimi and Kann(2021)}]{ebrahimi-kann-2021-adapt}
Abteen Ebrahimi and Katharina Kann. 2021.
\newblock \href {https://doi.org/10.18653/v1/2021.acl-long.351} {How to adapt your pretrained multilingual model to 1600 languages}.
\newblock In \emph{Proceedings of the 59th Annual Meeting of the Association for Computational Linguistics and the 11th International Joint Conference on Natural Language Processing (Volume 1: Long Papers)}, pages 4555--4567, Online. Association for Computational Linguistics.

\bibitem[{Faisal and Anastasopoulos(2022)}]{faisal-anastasopoulos-2022-phylogeny}
Fahim Faisal and Antonios Anastasopoulos. 2022.
\newblock \href {https://aclanthology.org/2022.aacl-main.34} {Phylogeny-inspired adaptation of multilingual models to new languages}.
\newblock In \emph{Proceedings of the 2nd Conference of the Asia-Pacific Chapter of the Association for Computational Linguistics and the 12th International Joint Conference on Natural Language Processing (Volume 1: Long Papers)}, pages 434--452, Online only. Association for Computational Linguistics.

\bibitem[{Freitag et~al.(2019)Freitag, Caswell, and Roy}]{freitag-etal-2019-ape}
Markus Freitag, Isaac Caswell, and Scott Roy. 2019.
\newblock \href {https://doi.org/10.18653/v1/W19-5204} {{APE} at scale and its implications on {MT} evaluation biases}.
\newblock In \emph{Proceedings of the Fourth Conference on Machine Translation (Volume 1: Research Papers)}, pages 34--44, Florence, Italy. Association for Computational Linguistics.

\bibitem[{Fujinuma et~al.(2022)Fujinuma, Boyd-Graber, and Kann}]{fujinuma-etal-2022-match}
Yoshinari Fujinuma, Jordan Boyd-Graber, and Katharina Kann. 2022.
\newblock \href {https://doi.org/10.18653/v1/2022.acl-long.106} {Match the script, adapt if multilingual: Analyzing the effect of multilingual pretraining on cross-lingual transferability}.
\newblock In \emph{Proceedings of the 60th Annual Meeting of the Association for Computational Linguistics (Volume 1: Long Papers)}, pages 1500--1512, Dublin, Ireland. Association for Computational Linguistics.

\bibitem[{Gellerstam(1986)}]{gellerstam1986translationese}
Martin Gellerstam. 1986.
\newblock Translationese in {S}wedish novels translated from {E}nglish.
\newblock \emph{Translation studies in Scandinavia}, 1:88--95.

\bibitem[{Gordon et~al.(2012)Gordon, Kozareva, and Roemmele}]{gordon-etal-2012-semeval}
Andrew Gordon, Zornitsa Kozareva, and Melissa Roemmele. 2012.
\newblock \href {https://aclanthology.org/S12-1052} {{S}em{E}val-2012 task 7: Choice of plausible alternatives: An evaluation of commonsense causal reasoning}.
\newblock In \emph{*{SEM} 2012: The First Joint Conference on Lexical and Computational Semantics {--} Volume 1: Proceedings of the main conference and the shared task, and Volume 2: Proceedings of the Sixth International Workshop on Semantic Evaluation ({S}em{E}val 2012)}, pages 394--398, Montr{\'e}al, Canada. Association for Computational Linguistics.

\bibitem[{He et~al.(2021)He, Liu, Ye, Tan, Ding, Cheng, Low, Bing, and Si}]{he-etal-2021-effectiveness}
Ruidan He, Linlin Liu, Hai Ye, Qingyu Tan, Bosheng Ding, Liying Cheng, Jiawei Low, Lidong Bing, and Luo Si. 2021.
\newblock \href {https://doi.org/10.18653/v1/2021.acl-long.172} {On the effectiveness of adapter-based tuning for pretrained language model adaptation}.
\newblock In \emph{Proceedings of the 59th Annual Meeting of the Association for Computational Linguistics and the 11th International Joint Conference on Natural Language Processing (Volume 1: Long Papers)}, pages 2208--2222, Online. Association for Computational Linguistics.

\bibitem[{Houlsby et~al.(2019)Houlsby, Giurgiu, Jastrzebski, Morrone, De~Laroussilhe, Gesmundo, Attariyan, and Gelly}]{pmlr-v97-houlsby19a}
Neil Houlsby, Andrei Giurgiu, Stanislaw Jastrzebski, Bruna Morrone, Quentin De~Laroussilhe, Andrea Gesmundo, Mona Attariyan, and Sylvain Gelly. 2019.
\newblock \href {https://proceedings.mlr.press/v97/houlsby19a.html} {Parameter-efficient transfer learning for {NLP}}.
\newblock In \emph{Proceedings of the 36th International Conference on Machine Learning}, volume~97 of \emph{Proceedings of Machine Learning Research}, pages 2790--2799. PMLR.

\bibitem[{Hu et~al.(2021)Hu, Shen, Wallis, Allen{-}Zhu, Li, Wang, and Chen}]{hu2021lora}
Edward~J. Hu, Yelong Shen, Phillip Wallis, Zeyuan Allen{-}Zhu, Yuanzhi Li, Shean Wang, and Weizhu Chen. 2021.
\newblock \href {http://arxiv.org/abs/2106.09685} {Lora: Low-rank adaptation of large language models}.
\newblock \emph{CoRR}, abs/2106.09685.

\bibitem[{Kew et~al.(2023)Kew, Schottmann, and Sennrich}]{kew2023turning}
Tannon Kew, Florian Schottmann, and Rico Sennrich. 2023.
\newblock \href {https://arxiv.org/pdf/2312.12683.pdf} {Turning english-centric llms into polyglots: How much multilinguality is needed?}
\newblock \emph{arXiv preprint arXiv:2312.12683}.

\bibitem[{Kim et~al.(2021)Kim, Shum, Susanj, and Hilgart}]{kim-etal-2021-revisiting}
Seungwon Kim, Alex Shum, Nathan Susanj, and Jonathan Hilgart. 2021.
\newblock \href {https://doi.org/10.18653/v1/2021.repl4nlp-1.11} {Revisiting pretraining with adapters}.
\newblock In \emph{Proceedings of the 6th Workshop on Representation Learning for NLP (RepL4NLP-2021)}, pages 90--99, Online. Association for Computational Linguistics.

\bibitem[{Lauscher et~al.(2020)Lauscher, Ravishankar, Vuli{\'c}, and Glava{\v{s}}}]{lauscher-etal-2020-zero}
Anne Lauscher, Vinit Ravishankar, Ivan Vuli{\'c}, and Goran Glava{\v{s}}. 2020.
\newblock \href {https://doi.org/10.18653/v1/2020.emnlp-main.363} {From zero to hero: {O}n the limitations of zero-shot language transfer with multilingual {T}ransformers}.
\newblock In \emph{Proceedings of the 2020 Conference on Empirical Methods in Natural Language Processing (EMNLP)}, pages 4483--4499, Online. Association for Computational Linguistics.

\bibitem[{McCloskey and Cohen(1989)}]{mccloskey1989catastrophic}
Michael McCloskey and Neal~J Cohen. 1989.
\newblock Catastrophic interference in connectionist networks: The sequential learning problem.
\newblock In \emph{Psychology of learning and motivation}, volume~24, pages 109--165. Elsevier.

\bibitem[{Pfeiffer et~al.(2022)Pfeiffer, Goyal, Lin, Li, Cross, Riedel, and Artetxe}]{pfeiffer-etal-2022-lifting}
Jonas Pfeiffer, Naman Goyal, Xi~Lin, Xian Li, James Cross, Sebastian Riedel, and Mikel Artetxe. 2022.
\newblock \href {https://doi.org/10.18653/v1/2022.naacl-main.255} {Lifting the curse of multilinguality by pre-training modular transformers}.
\newblock In \emph{Proceedings of the 2022 Conference of the North American Chapter of the Association for Computational Linguistics: Human Language Technologies}, pages 3479--3495, Seattle, United States. Association for Computational Linguistics.

\bibitem[{Pfeiffer et~al.(2020{\natexlab{a}})Pfeiffer, R{\"u}ckl{\'e}, Poth, Kamath, Vuli{\'c}, Ruder, Cho, and Gurevych}]{pfeiffer-etal-2020-adapterhub}
Jonas Pfeiffer, Andreas R{\"u}ckl{\'e}, Clifton Poth, Aishwarya Kamath, Ivan Vuli{\'c}, Sebastian Ruder, Kyunghyun Cho, and Iryna Gurevych. 2020{\natexlab{a}}.
\newblock \href {https://doi.org/10.18653/v1/2020.emnlp-demos.7} {{A}dapter{H}ub: A framework for adapting transformers}.
\newblock In \emph{Proceedings of the 2020 Conference on Empirical Methods in Natural Language Processing: System Demonstrations}, pages 46--54, Online. Association for Computational Linguistics.

\bibitem[{Pfeiffer et~al.(2023)Pfeiffer, Ruder, Vuli{\'c}, and Ponti}]{pfeiffer2023modular}
Jonas Pfeiffer, Sebastian Ruder, Ivan Vuli{\'c}, and Edoardo~Maria Ponti. 2023.
\newblock Modular deep learning.
\newblock \emph{arXiv preprint arXiv:2302.11529}.

\bibitem[{Pfeiffer et~al.(2020{\natexlab{b}})Pfeiffer, Vuli{\'c}, Gurevych, and Ruder}]{pfeiffer-etal-2020-mad}
Jonas Pfeiffer, Ivan Vuli{\'c}, Iryna Gurevych, and Sebastian Ruder. 2020{\natexlab{b}}.
\newblock \href {https://doi.org/10.18653/v1/2020.emnlp-main.617} {{MAD-X}: {A}n {A}dapter-{B}ased {F}ramework for {M}ulti-{T}ask {C}ross-{L}ingual {T}ransfer}.
\newblock In \emph{Proceedings of the 2020 Conference on Empirical Methods in Natural Language Processing (EMNLP)}, pages 7654--7673, Online. Association for Computational Linguistics.

\bibitem[{Ponti et~al.(2020)Ponti, Glava{\v{s}}, Majewska, Liu, Vuli{\'c}, and Korhonen}]{ponti-etal-2020-xcopa}
Edoardo~Maria Ponti, Goran Glava{\v{s}}, Olga Majewska, Qianchu Liu, Ivan Vuli{\'c}, and Anna Korhonen. 2020.
\newblock \href {https://doi.org/10.18653/v1/2020.emnlp-main.185} {{XCOPA}: A multilingual dataset for causal commonsense reasoning}.
\newblock In \emph{Proceedings of the 2020 Conference on Empirical Methods in Natural Language Processing (EMNLP)}, pages 2362--2376, Online. Association for Computational Linguistics.

\bibitem[{Ratcliff(1990)}]{ratcliff1990connectionist}
Roger Ratcliff. 1990.
\newblock Connectionist models of recognition memory: constraints imposed by learning and forgetting functions.
\newblock \emph{Psychological review}, 97(2):285.

\bibitem[{Roemmele et~al.(2011)Roemmele, Bejan, and Gordon}]{roemmele2011choice}
Melissa Roemmele, Cosmin~Adrian Bejan, and Andrew~S Gordon. 2011.
\newblock Choice of plausible alternatives: An evaluation of commonsense causal reasoning.
\newblock In \emph{AAAI spring symposium: logical formalizations of commonsense reasoning}, pages 90--95.

\bibitem[{R{\"u}ckl{\'e} et~al.(2021)R{\"u}ckl{\'e}, Geigle, Glockner, Beck, Pfeiffer, Reimers, and Gurevych}]{ruckle-etal-2021-adapterdrop}
Andreas R{\"u}ckl{\'e}, Gregor Geigle, Max Glockner, Tilman Beck, Jonas Pfeiffer, Nils Reimers, and Iryna Gurevych. 2021.
\newblock \href {https://doi.org/10.18653/v1/2021.emnlp-main.626} {{AdapterDrop}: {O}n the efficiency of adapters in transformers}.
\newblock In \emph{Proceedings of the 2021 Conference on Empirical Methods in Natural Language Processing}, pages 7930--7946, Online and Punta Cana, Dominican Republic. Association for Computational Linguistics.

\bibitem[{Sap et~al.(2019)Sap, Rashkin, Chen, Le~Bras, and Choi}]{sap-etal-2019-social}
Maarten Sap, Hannah Rashkin, Derek Chen, Ronan Le~Bras, and Yejin Choi. 2019.
\newblock \href {https://doi.org/10.18653/v1/D19-1454} {Social {IQ}a: Commonsense reasoning about social interactions}.
\newblock In \emph{Proceedings of the 2019 Conference on Empirical Methods in Natural Language Processing and the 9th International Joint Conference on Natural Language Processing (EMNLP-IJCNLP)}, pages 4463--4473, Hong Kong, China. Association for Computational Linguistics.

\bibitem[{Sinha et~al.(2021)Sinha, Parthasarathi, Pineau, and Williams}]{sinha-etal-2021-unnatural}
Koustuv Sinha, Prasanna Parthasarathi, Joelle Pineau, and Adina Williams. 2021.
\newblock \href {https://doi.org/10.18653/v1/2021.acl-long.569} {{UnNatural} {L}anguage {I}nference}.
\newblock In \emph{Proceedings of the 59th Annual Meeting of the Association for Computational Linguistics and the 11th International Joint Conference on Natural Language Processing (Volume 1: Long Papers)}, pages 7329--7346, Online. Association for Computational Linguistics.

\bibitem[{{\"U}st{\"u}n et~al.(2020){\"U}st{\"u}n, Bisazza, Bouma, and van Noord}]{ustun-etal-2020-udapter}
Ahmet {\"U}st{\"u}n, Arianna Bisazza, Gosse Bouma, and Gertjan van Noord. 2020.
\newblock \href {https://doi.org/10.18653/v1/2020.emnlp-main.180} {{UD}apter: Language adaptation for truly {U}niversal {D}ependency parsing}.
\newblock In \emph{Proceedings of the 2020 Conference on Empirical Methods in Natural Language Processing (EMNLP)}, pages 2302--2315, Online. Association for Computational Linguistics.

\bibitem[{Wang et~al.(2019)Wang, Pruksachatkun, Nangia, Singh, Michael, Hill, Levy, and Bowman}]{superglue}
Alex Wang, Yada Pruksachatkun, Nikita Nangia, Amanpreet Singh, Julian Michael, Felix Hill, Omer Levy, and Samuel Bowman. 2019.
\newblock \href {https://proceedings.neurips.cc/paper/2019/file/4496bf24afe7fab6f046bf4923da8de6-Paper.pdf} {Superglue: A stickier benchmark for general-purpose language understanding systems}.
\newblock In \emph{Advances in Neural Information Processing Systems}, volume~32. Curran Associates, Inc.

\bibitem[{Williams et~al.(2018)Williams, Nangia, and Bowman}]{williams-etal-2018-broad}
Adina Williams, Nikita Nangia, and Samuel Bowman. 2018.
\newblock \href {https://doi.org/10.18653/v1/N18-1101} {A broad-coverage challenge corpus for sentence understanding through inference}.
\newblock In \emph{Proceedings of the 2018 Conference of the North {A}merican Chapter of the Association for Computational Linguistics: Human Language Technologies, Volume 1 (Long Papers)}, pages 1112--1122, New Orleans, Louisiana. Association for Computational Linguistics.

\bibitem[{Yang et~al.(2019)Yang, Zhang, Tar, and Baldridge}]{yang-etal-2019-paws}
Yinfei Yang, Yuan Zhang, Chris Tar, and Jason Baldridge. 2019.
\newblock \href {https://doi.org/10.18653/v1/D19-1382} {{PAWS}-{X}: A cross-lingual adversarial dataset for paraphrase identification}.
\newblock In \emph{Proceedings of the 2019 Conference on Empirical Methods in Natural Language Processing and the 9th International Joint Conference on Natural Language Processing (EMNLP-IJCNLP)}, pages 3687--3692, Hong Kong, China. Association for Computational Linguistics.

\bibitem[{Yong et~al.(2022)Yong, Schoelkopf, Muennighoff, Aji, Adelani, Almubarak, Bari, Sutawika, Kasai, Baruwa et~al.}]{yong2022bloom}
Zheng-Xin Yong, Hailey Schoelkopf, Niklas Muennighoff, Alham~Fikri Aji, David~Ifeoluwa Adelani, Khalid Almubarak, M~Saiful Bari, Lintang Sutawika, Jungo Kasai, Ahmed Baruwa, et~al. 2022.
\newblock \href {https://arxiv.org/pdf/2212.09535} {Bloom+ 1: Adding language support to bloom for zero-shot prompting}.
\newblock \emph{arXiv preprint arXiv:2212.09535}.

\bibitem[{Zhang et~al.(2019)Zhang, Baldridge, and He}]{zhang-etal-2019-paws}
Yuan Zhang, Jason Baldridge, and Luheng He. 2019.
\newblock \href {https://doi.org/10.18653/v1/N19-1131} {{PAWS}: Paraphrase adversaries from word scrambling}.
\newblock In \emph{Proceedings of the 2019 Conference of the North {A}merican Chapter of the Association for Computational Linguistics: Human Language Technologies, Volume 1 (Long and Short Papers)}, pages 1298--1308, Minneapolis, Minnesota. Association for Computational Linguistics.

\end{thebibliography}

\appendix
\newpage
\section{Full results}
\label{appx:full_results}

In this section, we present the full results for both models, all three tasks, and all language pairs. 

\paragraph{XNLI. }For XNLI, we report the results for each source language in the following tables, in decreasing order of the languages' representation in the pre-training corpora of the models: English (Table \ref{tab:xnli_en_app}), German (Table \ref{tab:xnli_de_app}), Russian (Table \ref{tab:xnli_ru_app}), Spanish (Table \ref{tab:xnli_es_app}), Chinese (Table \ref{tab:xnli_zh_app}), Vietnamese (Table \ref{tab:xnli_vi_app}), Arabic (Table \ref{tab:xnli_ar_app}), Turkish (Table \ref{tab:xnli_tr_app}), Greek (Table \ref{tab:xnli_el_app}), Hindi (Table \ref{tab:xnli_hi_app}), and Swahili (Table \ref{tab:xnli_sw_app}). For XLM-R, note the better performance of the Target compared to the Source setup for source languages other than English, which we discussed in section \ref{secsec:rq3}. For mBERT however, the patterns for the other source languages are similar to the patterns for English. 

\paragraph{PAWS-X. }For PAWS-X, the results for each source language are found in the following tables, ordered from highest resource to lowest resource: English (Table \ref{tab:paws-x_en_app}), German (Table \ref{tab:paws-x_de_app}), Spanish (Table \ref{tab:paws-x_es_app}), Japanese (Table \ref{tab:paws-x_ja_app}), and Chinese (Table \ref{tab:paws-x_zh_app}). For this dataset, we do not observe major differences between different source languages. 

\paragraph{XCOPA. }Lastly, for XCOPA, there exists a training set only for English. Therefore, we cannot provide results for other source languages. The results for English are detailed in Table \ref{tab:xcopa_en_app}.

\paragraph{The impact of source language pre-training resources on the performance. }
Another observation we would like to draw attention to is the fact that we \textit{do not} observe a tendency that higher-resource source languages lead to a higher performance in cross-lingual transfer: For English as a source language, the best result for XLM-R and XNLI is $70.7\%$ and for mBERT and XNLI, it is $63.0\%$ accuracy. For the lowest-resource language, Swahili, the corresponding numbers are $72.2\%$ accuracy for XLM-R and $61.3\%$ accuracy for mBERT. For PAWS-X, for English, the best result for XLM-R is $78.2\%$; for mBERT, it is $77.2\%$. For the lowest-resource language Chinese, the corresponding numbers are higher: $81.9\%$ for XLM-R and $78.6\%$ for mBERT. While the increase is likely to be caused by the fact that the target languages for lower-resource languages are relatively higher-resourced, the patterns we observe show that the amount of pre-training resources of the source language is not of importance for these two datasets. 



\begin{table*}[t]
\centering
\begin{tabular}{ c c c c c c c c c} 
\toprule
\multicolumn{1}{c}{} & \multicolumn{4}{c}{XLM-R} &  \multicolumn{4}{c}{mBERT} \\
& Target & English & None & None$_{tr}$ & Target & English & None & None$_{tr}$ \\ \cmidrule(lr){2-5} \cmidrule(lr){6-9}
en & (81.8) & (81.8) & (81.5) & (81.7) & (78.1) & (78.1) & (70.9) & (77.7) \\
de & 73.6 & 73.3 & 73.4 & 73.6 & 66.1 & 67.9 & 58.1 & 67.5 \\
ru & 72.4 & 72.4 & 72.7 & 72.8 & 64.1 & 64.6 & 55.0 & 64.1 \\
es & 76.0 & 76.2 & 75.9 & 75.9 & 69.1 & 71.4 & 62.5 & 70.5 \\
zh & 70.0 & 71.7 & 70.8 & 71.0 & 66.3 & 67.4 & 57.7 & 65.8 \\
vi & 71.6 & 71.5 & 71.3 & 71.8 & 68.2 & 68.4 & 58.7 & 66.8 \\
ar & 68.6 & 65.8 & 68.2 & 68.8 & 38.7 & 62.7 & 50.7 & 61.9 \\
tr & 69.8 & 70.7 & 70.2 & 71.0 & 62.0 & 61.3 & 50.6 & 60.4 \\
el & 72.3 & 71.9 & 71.8 & 72.0 & 60.8 & 60.9 & 54.0 & 60.2 \\
hi & 66.7 & 67.1 & 66.9 & 67.2 & 57.1 & 57.4 & 47.6 & 56.5 \\
sw & 65.2 & 59.0 & 62.4 & 62.7 & 37.4 & 47.7 & 40.8 & 48.2 \\
\midrule Avg. & 70.6 & 70.0 & 70.4 & 70.7 & 59.0 & 63.0 & 53.6 & 62.2 \\ \bottomrule
\end{tabular}
\caption{Results on XNLI with transfer from English (en) into all evaluated target languages, ordered by pre-training resources top-to-bottom. Results on English are included for reference but excluded from the average.}
\label{tab:xnli_en_app}
\end{table*}

\begin{table*}[t]
\centering
\begin{tabular}{ c c c c c c c c c} 
\toprule
\multicolumn{1}{c}{} & \multicolumn{4}{c}{XLM-R} &  \multicolumn{4}{c}{mBERT} \\
& Target & German & None & None$_{tr}$ & Target & German & None & None$_{tr}$ \\ \cmidrule(lr){2-5} \cmidrule(lr){6-9}
en & 80.0 & 78.7 & 79.1 & 80.5 & 74.3 & 74.2 & 67.9 & 74.2 \\
de & (76.1) & (76.1) & (74.9) & (75.6) & (71.9) & (71.9) & (65.9) & (71.2) \\
ru & 73.5 & 71.4 & 72.7 & 74.1 & 66.6 & 66.5 & 59.7 & 66.0 \\
es & 76.4 & 74.1 & 75.0 & 76.5 & 71.5 & 71.6 & 64.7 & 70.9 \\
zh & 73.4 & 72.7 & 72.9 & 73.8 & 67.6 & 68.4 & 60.1 & 67.4 \\
vi & 73.5 & 71.3 & 72.1 & 73.4 & 67.3 & 68.0 & 60.2 & 67.3 \\
ar & 70.6 & 63.4 & 69.4 & 71.1 & 42.0 & 62.4 & 53.2 & 63.7 \\
tr & 71.6 & 67.4 & 70.9 & 72.9 & 62.8 & 60.9 & 53.2 & 61.4 \\
el & 73.1 & 69.0 & 72.2 & 73.1 & 61.6 & 62.1 & 55.7 & 61.8 \\
hi & 68.8 & 65.9 & 68.5 & 69.6 & 58.4 & 58.0 & 50.1 & 58.8 \\
sw & 66.7 & 51.8 & 63.1 & 64.2 & 36.5 & 45.7 & 40.3 & 49.3 \\
\midrule Avg. & 72.8 & 68.6 & 71.6 & 72.9 & 60.9 & 63.8 & 56.5 & 64.1 \\ \bottomrule
\end{tabular}
\caption{Results on XNLI with transfer from German (de) into all evaluated target languages, ordered by pre-training resources top-to-bottom. Results on German are included for reference but excluded from the average.}
\label{tab:xnli_de_app}
\end{table*}

\begin{table*}[t]
\centering
\begin{tabular}{ c c c c c c c c c} 
\toprule
\multicolumn{1}{c}{} & \multicolumn{4}{c}{XLM-R} &  \multicolumn{4}{c}{mBERT} \\
& Target & Russian & None & None$_{tr}$ & Target & Russian & None & None$_{tr}$ \\ \cmidrule(lr){2-5} \cmidrule(lr){6-9}
en & 80.3 & 79.4 & 79.8 & 80.7 & 73.3 & 73.2 & 69.0 & 73.5 \\
de & 74.5 & 73.3 & 73.8 & 74.9 & 67.3 & 68.5 & 63.3 & 68.7 \\
ru & (74.7) & (74.7) & (74.0) & (74.9) & (69.5) & (69.5) & (64.2) & (69.4) \\
es & 76.1 & 75.1 & 75.8 & 76.7 & 70.6 & 70.8 & 66.2 & 70.8 \\
zh & 73.3 & 73.1 & 72.6 & 73.3 & 66.7 & 68.0 & 61.0 & 67.7 \\
vi & 73.4 & 73.7 & 72.5 & 73.8 & 66.9 & 65.7 & 62.0 & 67.7 \\
ar & 70.3 & 67.0 & 69.6 & 71.2 & 38.9 & 57.9 & 56.6 & 63.0 \\
tr & 71.5 & 68.1 & 71.2 & 72.2 & 62.4 & 54.4 & 56.3 & 61.0 \\
el & 73.3 & 70.0 & 72.9 & 73.8 & 60.5 & 58.4 & 58.0 & 61.9 \\
hi & 69.4 & 67.0 & 68.9 & 69.6 & 56.5 & 53.3 & 52.1 & 59.1 \\
sw & 67.8 & 56.7 & 64.6 & 64.5 & 40.2 & 39.0 & 44.2 & 47.2 \\
\midrule Avg. & 73.0 & 70.3 & 72.2 & 73.1 & 60.3 & 60.9 & 58.9 & 64.1 \\ \bottomrule
\end{tabular}
\caption{Results on XNLI with transfer from Russian (ru) into all evaluated target languages, ordered by pre-training resources top-to-bottom. Results on Russian are included for reference but excluded from the average.}
\label{tab:xnli_ru_app}
\end{table*}

\begin{table*}[t]
\centering
\begin{tabular}{ c c c c c c c c c} 
\toprule
\multicolumn{1}{c}{} & \multicolumn{4}{c}{XLM-R} &  \multicolumn{4}{c}{mBERT} \\
& Target & Spanish & None & None$_{tr}$ & Target & Spanish & None & None$_{tr}$ \\ \cmidrule(lr){2-5} \cmidrule(lr){6-9}
en & 80.2 & 79.5 & 79.5 & 80.5 & 75.4 & 75.3 & 71.7 & 75.0 \\
de & 74.0 & 71.7 & 73.4 & 74.8 & 69.0 & 68.0 & 65.2 & 68.4 \\
ru & 72.7 & 71.5 & 71.9 & 73.7 & 66.5 & 66.5 & 61.9 & 65.3 \\
es & (76.9) & (76.9) & (75.9) & (77.1) & (74.2) & (74.2) & (70.2) & (73.9) \\
zh & 71.4 & 71.7 & 71.2 & 73.0 & 67.1 & 68.6 & 63.0 & 67.4 \\
vi & 72.3 & 72.0 & 71.6 & 73.6 & 66.1 & 68.3 & 63.4 & 67.5 \\
ar & 67.2 & 67.8 & 67.7 & 70.4 & 42.6 & 62.7 & 57.2 & 62.7 \\
tr & 70.6 & 66.8 & 70.2 & 71.9 & 60.7 & 59.1 & 55.3 & 60.3 \\
el & 72.1 & 69.9 & 71.4 & 73.1 & 62.0 & 61.7 & 58.1 & 61.5 \\
hi & 67.7 & 66.1 & 67.6 & 69.1 & 57.2 & 56.4 & 51.9 & 57.6 \\
sw & 65.6 & 55.5 & 62.6 & 63.2 & 38.1 & 45.0 & 45.8 & 48.3 \\
\midrule Avg. & 71.4 & 69.2 & 70.7 & 72.3 & 60.5 & 63.2 & 59.4 & 63.4  \\ \bottomrule
\end{tabular}
\caption{Results on XNLI with transfer from Spanish (es) into all evaluated target languages, ordered by pre-training resources top-to-bottom. Results on Spanish are included for reference but excluded from the average.}
\label{tab:xnli_es_app}
\end{table*}

\begin{table*}[t]
\centering
\begin{tabular}{ c c c c c c c c c} 
\toprule
\multicolumn{1}{c}{} & \multicolumn{4}{c}{XLM-R} &  \multicolumn{4}{c}{mBERT} \\
& Target & Chinese & None & None$_{tr}$ & Target & Chinese & None & None$_{tr}$ \\ \cmidrule(lr){2-5} \cmidrule(lr){6-9}
en & 78.7 & 78.0 & 77.8 & 79.0 & 73.4 & 73.1 & 70.9 & 72.6 \\
de & 72.9 & 71.8 & 71.4 & 73.7 & 66.2 & 67.6 & 65.2 & 67.1 \\
ru & 72.3 & 69.0 & 70.8 & 72.6 & 65.1 & 65.6 & 63.4 & 66.0 \\
es & 74.6 & 73.9 & 73.5 & 75.5 & 69.0 & 70.1 & 67.9 & 69.6 \\
zh & (73.7) & (73.7) & (72.7) & (74.4) & (72.1) & (72.1) & (68.9) & (71.5) \\
vi & 72.5 & 73.2 & 71.4 & 73.5 & 66.9 & 68.5 & 64.8 & 67.7 \\
ar & 68.9 & 65.2 & 67.6 & 69.9 & 34.7 & 62.5 & 59.6 & 62.3 \\
tr & 69.6 & 65.3 & 69.4 & 71.7 & 61.9 & 59.2 & 58.2 & 60.7 \\
el & 71.0 & 69.2 & 70.5 & 72.5 & 58.3 & 60.4 & 58.8 & 60.5 \\
hi & 67.3 & 64.0 & 66.8 & 68.8 & 57.2 & 58.3 & 54.2 & 58.9 \\
sw & 65.6 & 50.6 & 62.6 & 64.0 & 33.7 & 42.4 & 44.9 & 43.7 \\
\midrule Avg. & 71.3 & 68.0 & 70.2 & 72.1 & 58.6 & 62.8 & 60.8 & 62.9 \\ \bottomrule
\end{tabular}
\caption{Results on XNLI with transfer from Chinese (zh) into all evaluated target languages, ordered by pre-training resources top-to-bottom. Results on Chinese are included for reference but excluded from the average.}
\label{tab:xnli_zh_app}
\end{table*}

\begin{table*}[t]
\centering
\begin{tabular}{ c c c c c c c c c} 
\toprule
\multicolumn{1}{c}{} & \multicolumn{4}{c}{XLM-R} &  \multicolumn{4}{c}{mBERT} \\
& Target & Vietnamese & None & None$_{tr}$ & Target & Vietnamese & None & None$_{tr}$ \\ \cmidrule(lr){2-5} \cmidrule(lr){6-9}
en & 78.3 & 77.1 & 76.9 & 79.5 & 72.6 & 71.8 & 70.0 & 72.3 \\
de & 73.6 & 69.4 & 71.0 & 74.2 & 66.8 & 66.8 & 64.4 & 66.4 \\
ru & 72.6 & 69.2 & 69.1 & 73.5 & 65.4 & 64.7 & 61.9 & 64.8 \\
es & 75.3 & 72.0 & 72.4 & 75.9 & 69.2 & 70.1 & 67.4 & 69.5 \\
zh & 72.5 & 71.0 & 70.1 & 73.3 & 66.3 & 69.1 & 65.9 & 68.0 \\
vi & (74.7) & (74.7) & (70.9) & (74.8) & (71.0) & (71.0) & (68.5) & (70.3) \\
ar & 69.9 & 63.0 & 67.2 & 70.4 & 39.5 & 61.0 & 58.5 & 62.0 \\
tr & 71.8 & 70.0 & 68.4 & 72.3 & 63.4 & 60.3 & 59.3 & 60.1 \\
el & 72.7 & 65.1 & 69.9 & 73.1 & 60.8 & 61.1 & 60.6 & 61.9 \\
hi & 68.9 & 63.8 & 66.8 & 69.1 & 58.5 & 58.1 & 55.8 & 57.8 \\
sw & 65.7 & 50.4 & 61.1 & 63.5 & 37.8 & 46.4 & 47.1 & 48.6 \\
\midrule Avg. & 72.1 & 67.1 & 69.3 & 72.5 & 60.0 & 62.9 & 61.1 & 63.1 \\ \bottomrule
\end{tabular}
\caption{Results on XNLI with transfer from Vietnamese (vi) into all evaluated target languages, ordered by pre-training resources top-to-bottom. Results on Vietnamese are included for reference but excluded from the average.}
\label{tab:xnli_vi_app}
\end{table*}

\begin{table*}[t]
\centering
\begin{tabular}{ c c c c c c c c c} 
\toprule
\multicolumn{1}{c}{} & \multicolumn{4}{c}{XLM-R} &  \multicolumn{4}{c}{mBERT} \\
& Target & Arabic & None & None$_{tr}$ & Target & Arabic & None & None$_{tr}$ \\ \cmidrule(lr){2-5} \cmidrule(lr){6-9}
en & 78.4 & 78.4 & 76.5 & 79.9 & 69.6 & 71.4 & 63.4 & 71.4 \\
de & 72.5 & 73.8 & 69.8 & 74.5 & 65.2 & 66.8 & 60.0 & 66.5 \\
ru & 71.4 & 68.8 & 68.1 & 73.4 & 62.5 & 64.4 & 57.0 & 64.0 \\
es & 75.0 & 75.1 & 72.8 & 76.3 & 67.1 & 69.7 & 61.8 & 69.9 \\
zh & 71.0 & 72.1 & 68.0 & 72.9 & 65.1 & 67.3 & 60.7 & 66.5 \\
vi & 72.3 & 73.1 & 69.0 & 73.4 & 64.5 & 63.3 & 58.8 & 66.8 \\
ar & (72.6) & (72.6) & (68.7) & (72.3) & (67.1) & (67.1) & (59.5) & (65.9) \\
tr & 70.2 & 56.8 & 66.6 & 72.1 & 58.4 & 59.2 & 54.3 & 60.0 \\
el & 71.6 & 71.1 & 69.8 & 73.2 & 58.1 & 61.9 & 56.4 & 61.2 \\
hi & 67.4 & 67.7 & 65.0 & 68.8 & 57.2 & 57.8 & 53.0 & 56.6 \\
sw & 66.0 & 53.4 & 61.1 & 63.8 & 57.5 & 47.0 & 44.8 & 49.0 \\
\midrule Avg. & 71.6 & 69.0 & 68.7 & 72.8 & 62.5 & 62.9 & 57.0 & 63.2 \\ \bottomrule
\end{tabular}
\caption{Results on XNLI with transfer from Arabic (ar) into all evaluated target languages, ordered by pre-training resources top-to-bottom. Results on Arabic are included for reference but excluded from the average.}
\label{tab:xnli_ar_app}
\end{table*}

\begin{table*}[t]
\centering
\begin{tabular}{ c c c c c c c c c} 
\toprule
\multicolumn{1}{c}{} & \multicolumn{4}{c}{XLM-R} &  \multicolumn{4}{c}{mBERT} \\
& Target & Turkish & None & None$_{tr}$ & Target & Turkish & None & None$_{tr}$ \\ \cmidrule(lr){2-5} \cmidrule(lr){6-9}
en & 78.1 & 76.8 & 75.8 & 79.0 & 70.8 & 68.6 & 68.3 & 67.9 \\
de & 73.5 & 71.3 & 69.6 & 73.8 & 66.2 & 66.0 & 64.9 & 65.4 \\
ru & 72.4 & 70.6 & 67.6 & 73.4 & 64.1 & 63.9 & 61.8 & 62.4 \\
es & 74.8 & 72.7 & 71.2 & 75.7 & 66.8 & 67.2 & 65.8 & 66.6 \\
zh & 70.2 & 72.2 & 65.4 & 73.3 & 64.4 & 66.1 & 63.1 & 65.2 \\
vi & 72.3 & 71.1 & 66.7 & 73.0 & 65.8 & 65.5 & 62.7 & 65.1 \\
ar & 70.4 & 61.0 & 64.5 & 69.7 & 39.8 & 61.1 & 58.9 & 61.0 \\
tr & (73.7) & (73.7) & (68.0) & (73.7) & (68.0) & (68.0) & (64.5) & (67.1) \\
el & 71.8 & 68.1 & 68.2 & 72.3 & 59.9 & 59.3 & 59.2 & 59.9 \\
hi & 68.5 & 66.1 & 63.8 & 69.3 & 58.0 & 58.1 & 55.2 & 57.6 \\
sw & 66.2 & 53.1 & 58.4 & 64.8 & 36.3 & 48.2 & 47.2 & 50.4 \\
\midrule Avg. & 71.8 & 68.3 & 67.1 & 72.4 & 59.2 & 62.4 & 60.7 & 62.2 \\ \bottomrule
\end{tabular}
\caption{Results on XNLI with transfer from Turkish (tr) into all evaluated target languages, ordered by pre-training resources top-to-bottom. Results on Turkish are included for reference but excluded from the average.}
\label{tab:xnli_tr_app}
\end{table*}

\begin{table*}[t]
\centering
\begin{tabular}{ c c c c c c c c c} 
\toprule
\multicolumn{1}{c}{} & \multicolumn{4}{c}{XLM-R} &  \multicolumn{4}{c}{mBERT} \\
& Target & Greek & None & None$_{tr}$ & Target & Greek & None & None$_{tr}$ \\ \cmidrule(lr){2-5} \cmidrule(lr){6-9}
en & 79.5 & 78.7 & 78.4 & 79.9 & 69.3 & 68.9 & 64.7 & 70.6 \\
de & 74.6 & 73.2 & 73.7 & 74.7 & 66.0 & 66.1 & 62.1 & 66.3 \\
ru & 73.2 & 71.9 & 72.1 & 73.7 & 64.2 & 63.5 & 60.3 & 64.8 \\
es & 76.5 & 75.4 & 75.5 & 76.5 & 67.9 & 68.3 & 64.3 & 69.0 \\
zh & 72.2 & 71.1 & 71.5 & 73.4 & 60.0 & 65.0 & 60.4 & 65.3 \\
vi & 72.6 & 72.8 & 71.3 & 73.3 & 64.5 & 64.2 & 61.8 & 65.4 \\
ar & 69.9 & 68.6 & 69.3 & 70.9 & 45.7 & 59.0 & 57.3 & 61.7 \\
tr & 70.7 & 67.8 & 69.8 & 71.8 & 60.5 & 57.9 & 55.9 & 60.5 \\
el & (74.4) & (74.4) & (73.2) & (73.8) & (65.9) & (65.9) & (61.2) & (64.8) \\
hi & 68.3 & 66.0 & 67.8 & 69.2 & 55.6 & 54.0 & 52.2 & 57.9 \\
sw & 67.0 & 58.6 & 63.1 & 64.5 & 41.0 & 43.3 & 45.4 & 49.2 \\
\midrule Avg. & 72.5 & 70.4 & 71.2 & 72.8 & 59.5 & 61.0 & 58.4 & 63.1 \\ \bottomrule
\end{tabular}
\caption{Results on XNLI with transfer from Greek (el) into all evaluated target languages, ordered by pre-training resources top-to-bottom. Results on Greek are included for reference but excluded from the average.}
\label{tab:xnli_el_app}
\end{table*}

\begin{table*}[t]
\centering
\begin{tabular}{ c c c c c c c c c} 
\toprule
\multicolumn{1}{c}{} & \multicolumn{4}{c}{XLM-R} &  \multicolumn{4}{c}{mBERT} \\
& Target & Hindi & None & None$_{tr}$ & Target & Hindi & None & None$_{tr}$ \\ \cmidrule(lr){2-5} \cmidrule(lr){6-9}
en & 77.7 & 76.3 & 76.6 & 77.3 & 68.0 & 66.7 & 61.7 & 68.4 \\
de & 72.7 & 69.1 & 70.4 & 72.5 & 64.5 & 64.4 & 61.1 & 64.7 \\
ru & 71.7 & 68.3 & 69.0 & 71.8 & 62.8 & 62.5 & 58.8 & 63.9 \\
es & 73.9 & 70.4 & 71.8 & 73.6 & 66.0 & 66.0 & 62.2 & 65.3 \\
zh & 70.7 & 67.8 & 68.2 & 71.2 & 65.8 & 65.4 & 61.7 & 64.8 \\
vi & 71.8 & 71.4 & 69.8 & 71.6 & 65.9 & 64.9 & 61.2 & 65.3 \\
ar & 69.0 & 63.6 & 66.3 & 69.1 & 36.9 & 58.2 & 56.3 & 60.8 \\
tr & 70.9 & 65.3 & 68.6 & 70.9 & 62.0 & 58.2 & 57.4 & 60.6 \\
el & 71.5 & 66.7 & 70.1 & 71.4 & 60.4 & 57.7 & 58.3 & 60.6 \\
hi & (68.5) & (68.5) & (66.1) & (68.2) & (63.2) & (63.2) & (59.5) & (61.7) \\
sw & 66.3 & 56.0 & 61.1 & 63.1 & 33.9 & 46.9 & 46.5 & 50.1 \\
\midrule Avg. & 71.6 & 67.5 & 69.2 & 71.2 & 58.6 & 61.1 & 58.5 & 62.4 \\ \bottomrule
\end{tabular}
\caption{Results on XNLI with transfer from Hindi (hi) into all evaluated target languages, ordered by pre-training resources top-to-bottom. Results on Hindi are included for reference but excluded from the average.}
\label{tab:xnli_hi_app}
\end{table*}

\begin{table*}[t]
\centering
\begin{tabular}{ c c c c c c c c c} 
\toprule
\multicolumn{1}{c}{} & \multicolumn{4}{c}{XLM-R} &  \multicolumn{4}{c}{mBERT} \\
& Target & Swahili & None & None$_{tr}$ & Target & Swahili & None & None$_{tr}$ \\ \cmidrule(lr){2-5} \cmidrule(lr){6-9}
en & 78.1 & 77.6 & 77.2 & 77.3 & 67.6 & 69.5 & 53.5 & 67.4 \\
de & 73.0 & 70.7 & 72.1 & 72.1 & 56.6 & 62.8 & 47.4 & 59.6 \\
ru & 72.6 & 70.9 & 71.1 & 71.7 & 58.0 & 62.7 & 46.4 & 61.0 \\
es & 74.8 & 72.1 & 73.5 & 73.6 & 59.7 & 63.5 & 49.0 & 63.2 \\
zh & 71.8 & 70.5 & 70.7 & 72.1 & 60.8 & 63.6 & 44.9 & 61.7 \\
vi & 71.8 & 71.4 & 70.5 & 72.4 & 55.4 & 64.5 & 48.7 & 63.0 \\
ar & 68.6 & 66.7 & 67.9 & 69.5 & 60.7 & 58.7 & 42.8 & 59.0 \\
tr & 71.1 & 65.6 & 70.1 & 70.2 & 50.4 & 55.0 & 43.3 & 55.2 \\
el & 71.8 & 66.9 & 70.8 & 70.8 & 48.7 & 57.8 & 44.3 & 57.1 \\
hi & 68.0 & 65.0 & 67.3 & 68.0 & 49.5 & 55.1 & 42.1 & 52.9 \\
sw & (68.0) & (68.0) & (64.6) & (66.7) & (62.3) & (62.3) & (45.6) & (60.2) \\
\midrule Avg. & 72.2 & 69.7 & 71.1 & 71.8 & 56.7 & 61.3 & 46.2 & 60.0 \\ \bottomrule
\end{tabular}
\caption{Results on XNLI with transfer from Swahili (sw) into all evaluated target languages, ordered by pre-training resources top-to-bottom. Results on Swahili are included for reference but excluded from the average.}
\label{tab:xnli_sw_app}
\end{table*}

\begin{table*}[t]
\centering
\begin{tabular}{ c c c c c c c c c} 
\toprule
\multicolumn{1}{c}{} & \multicolumn{4}{c}{XLM-R} &  \multicolumn{4}{c}{mBERT} \\
& Target & English & None & None$_{tr}$ & Target & English & None & None$_{tr}$ \\ \cmidrule(lr){2-5} \cmidrule(lr){6-9}
en & (91.4) & (91.4) & (91.0) & (91.1) & (91.3) & (91.3) & (82.7) & (90.4) \\
de & 83.3 & 82.3 & 82.4 & 83.2 & 81.1 & 82.2 & 73.1 & 81.2 \\
es & 84.0 & 84.1 & 83.5 & 84.1 & 82.0 & 83.1 & 72.8 & 81.6 \\
ja & 69.7 & 69.2 & 69.6 & 70.2 & 69.7 & 69.9 & 64.1 & 69.1 \\
zh & 74.3 & 73.7 & 73.8 & 75.1 & 72.6 & 73.6 & 67.8 & 73.4 \\
\midrule Avg. & 77.8 & 77.3 & 77.3 & 78.2 & 76.4 & 77.2 & 69.4 & 76.3 \\ \bottomrule
\end{tabular}
\caption{Results on PAWS-X with transfer from English (en) into all evaluated target languages, ordered by pre-training resources top-to-bottom. Results on English are included for reference but excluded from the average.}
\label{tab:paws-x_en_app}
\end{table*}

\begin{table*}[t]
\centering
\begin{tabular}{ c c c c c c c c c} 
\toprule
\multicolumn{1}{c}{} & \multicolumn{4}{c}{XLM-R} &  \multicolumn{4}{c}{mBERT} \\
& Target & German & None & None$_{tr}$ & Target & German & None & None$_{tr}$ \\ \cmidrule(lr){2-5} \cmidrule(lr){6-9}
en & 90.1 & 89.3 & 89.4 & 89.8 & 86.9 & 87.8 & 80.7 & 86.2 \\
de & (84.5) & (84.5) & (83.9) & (84.3) & (81.6) & (81.6) & (74.3) & (81.0) \\
es & 84.3 & 83.6 & 83.7 & 84.2 & 78.9 & 80.8 & 74.3 & 79.8 \\
ja & 71.0 & 69.4 & 70.6 & 71.6 & 66.4 & 68.4 & 64.0 & 68.9 \\
zh & 75.2 & 74.2 & 75.0 & 75.1 & 71.7 & 73.1 & 68.8 & 72.0 \\
\midrule Avg. & 80.1 & 79.1 & 79.7 & 80.2 & 76.0 & 77.5 & 72.0 & 76.7 \\ \bottomrule
\end{tabular}
\caption{Results on PAWS-X with transfer from German (de) into all evaluated target languages, ordered by pre-training resources top-to-bottom. Results on German are included for reference but excluded from the average.}
\label{tab:paws-x_de_app}
\end{table*}

\begin{table*}[t]
\centering
\begin{tabular}{ c c c c c c c c c} 
\toprule
\multicolumn{1}{c}{} & \multicolumn{4}{c}{XLM-R} &  \multicolumn{4}{c}{mBERT} \\
& Target & Spanish & None & None$_{tr}$ & Target & Spanish & None & None$_{tr}$ \\ \cmidrule(lr){2-5} \cmidrule(lr){6-9}
en & 90.1 & 89.6 & 89.6 & 89.9 & 88.1 & 87.7 & 77.9 & 87.2 \\
de & 83.5 & 82.1 & 82.4 & 82.9 & 80.3 & 80.7 & 68.5 & 80.5 \\
es & (86.4) & (86.4) & (84.4) & (85.0) & (83.0) & (83.0) & (67.6) & (83.1) \\
ja & 70.9 & 67.7 & 69.4 & 70.4 & 67.3 & 69.2 & 62.2 & 69.5 \\
zh & 75.4 & 73.0 & 74.6 & 75.0 & 71.8 & 72.8 & 63.9 & 72.6 \\
\midrule Avg. & 80.0 & 78.1 & 79.0 & 79.6 & 76.9 & 77.6 & 68.1 & 77.4 \\ \bottomrule
\end{tabular}
\caption{Results on PAWS-X with transfer from Spanish (es) into all evaluated target languages, ordered by pre-training resources top-to-bottom. Results on Spanish are included for reference but excluded from the average.}
\label{tab:paws-x_es_app}
\end{table*}

\begin{table*}[t]
\centering
\begin{tabular}{ c c c c c c c c c} 
\toprule
\multicolumn{1}{c}{} & \multicolumn{4}{c}{XLM-R} &  \multicolumn{4}{c}{mBERT} \\
& Target & Japanese & None & None$_{tr}$ & Target & Japanese & None & None$_{tr}$ \\ \cmidrule(lr){2-5} \cmidrule(lr){6-9}
en & 87.3 & 87.0 & 86.9 & 87.2 & 74.9 & 78.0 & 73.1 & 75.4 \\
de & 82.0 & 80.8 & 81.4 & 81.7 & 72.3 & 74.4 & 70.7 & 71.7 \\
es & 81.4 & 80.2 & 80.9 & 82.7 & 72.2 & 75.7 & 71.7 & 73.2 \\
ja & (74.3) & (74.3) & (73.5) & (73.7) & (72.1) & (72.1) & (68.8) & (71.5) \\
zh & 77.3 & 77.0 & 77.4 & 77.1 & 73.5 & 74.1 & 69.7 & 72.6 \\
\midrule Avg. & 82.0 & 81.2 & 81.6 & 82.2 & 73.2 & 75.6 & 71.3 & 73.2 \\ \bottomrule
\end{tabular}
\caption{Results on PAWS-X with transfer from Japanese (ja) into all evaluated target languages, ordered by pre-training resources top-to-bottom. Results on Japanese are included for reference but excluded from the average.}
\label{tab:paws-x_ja_app}
\end{table*}

\begin{table*}[t]
\centering
\begin{tabular}{ c c c c c c c c c} 
\toprule
\multicolumn{1}{c}{} & \multicolumn{4}{c}{XLM-R} &  \multicolumn{4}{c}{mBERT} \\
& Target & Chinese & None & None$_{tr}$ & Target & Chinese & None & None$_{tr}$ \\ \cmidrule(lr){2-5} \cmidrule(lr){6-9}
en & 88.7 & 87.7 & 88.3 & 88.7 & 80.7 & 83.1 & 77.2 & 81.7 \\
de & 82.6 & 81.1 & 81.9 & 82.2 & 76.0 & 79.0 & 72.7 & 76.9 \\
es & 82.3 & 82.7 & 82.5 & 83.6 & 76.5 & 79.9 & 74.7 & 78.2 \\
ja & 73.2 & 72.4 & 72.8 & 73.1 & 71.2 & 72.4 & 67.6 & 71.4 \\
zh & (78.4) & (78.4) & (78.0) & (78.0) & (76.1) & (76.1) & (72.4) & (75.6) \\
\midrule Avg. & 81.7 & 81.0 & 81.4 & 81.9 & 76.1 & 78.6 & 73.1 & 77.1 \\ \bottomrule
\end{tabular}
\caption{Results on PAWS-X with transfer from Chinese (zh) into all evaluated target languages, ordered by pre-training resources top-to-bottom. Results on Chinese are included for reference but excluded from the average.}
\label{tab:paws-x_zh_app}
\end{table*}

\begin{table*}[t]
\centering
\begin{tabular}{ c c c c c c c c c} 
\toprule
\multicolumn{1}{c}{} & \multicolumn{4}{c}{XLM-R} &  \multicolumn{4}{c}{mBERT} \\
& Target & English & None & None$_{tr}$ & Target & English & None & None$_{tr}$ \\ \cmidrule(lr){2-5} \cmidrule(lr){6-9}
zh & 55.2 & 55.0 & 54.3 & 49.4 & 53.7 & 52.7 & 54.2 & 53.2 \\
vi & 55.3 & 54.9 & 55.1 & 52.8 & 51.6 & 52.9 & 51.1 & 52.6 \\
tr & 53.1 & 51.9 & 51.2 & 49.3 & 51.9 & 53.2 & 54.1 & 55.6 \\
id & 55.7 & 53.6 & 53.4 & 49.8 & 50.4 & 50.8 & 50.8 & 50.8 \\
et & 54.1 & 50.7 & 52.3 & 51.4 & 53.8 & 49.3 & 49.1 & 51.2 \\
sw & 54.0 & 49.7 & 52.0 & 49.7 & 50.0 & 50.4 & 50.5 & 49.1 \\
ht & 51.2 & 48.6 & 50.6 & 49.6 & 54.6 & 52.7 & 51.2 & 50.2 \\
qu & 51.4 & 51.2 & 49.6 & 50.2 & 52.6 & 48.5 & 49.8 & 48.2 \\
\midrule Avg. & 53.8 & 52.0 & 52.3 & 50.3 & 52.3 & 51.3 & 51.4 & 51.4 \\ \bottomrule
\end{tabular}
\caption{Results on XCOPA with transfer from English (en) into all evaluated target languages, ordered by pre-training resources top-to-bottom.}
\label{tab:xcopa_en_app}
\end{table*}

\end{document}